\documentclass{article}

\usepackage[preprint,nonatbib]{neurips_2018}

\usepackage[utf8]{inputenc} % allow utf-8 input
\usepackage[T1]{fontenc}    % use 8-bit T1 fonts
\usepackage{hyperref}       % hyperlinks
\usepackage{url}            % simple URL typesetting
\usepackage{booktabs}       % professional-quality tables
\usepackage{amsfonts}       % blackboard math symbols
\usepackage{nicefrac}       % compact symbols for 1/2, etc.
\usepackage{microtype}      % microtypography
\usepackage{graphicx} % Required for including images
\usepackage[ruled,vlined]{algorithm2e} % for the algorithms
\usepackage{amsmath} % Include the amsmath package
\usepackage{mathdots} % for diagonal dots in matrix
\usepackage[title]{appendix}
\usepackage[colorinlistoftodos]{todonotes}
\usepackage{todonotes} % for adding to do notes
\usepackage{multirow}

\title{LiPCoT: \underline{Li}near \underline{P}redictive \underline{Co}ding based \underline{T}okenizer for Self-supervised Learning of Time Series Data via Language Models}

\author{%
  Md Fahim~Anjum\thanks{Alternate email: dr.fahim.anjum@gmail.com} \\
  Department of Neurology\\
  University of California San Francisco\\
  San Francisco, CA 94143 \\
  \texttt{fahim.anjum@ucsf.edu} \\
}

\begin{document}

\maketitle

\begin{abstract}
  Language models have achieved remarkable success in various natural language processing tasks. However, their application to time series data, a crucial component in many domains, remains limited. This paper proposes LiPCoT (Linear Predictive Coding based Tokenizer for time series), a novel tokenizer that encodes time series data into a sequence of tokens, enabling self-supervised learning of time series using existing Language model architectures such as BERT. Unlike traditional time series tokenizers that rely heavily on CNN encoder for time series feature generation, LiPCoT employs stochastic modeling through linear predictive coding to create a latent space for time series providing a compact yet rich representation of the inherent stochastic nature of the data. Furthermore, LiPCoT is computationally efficient and can effectively handle time series data with varying sampling rates and lengths, overcoming common limitations of existing time series tokenizers. In this proof-of-concept work, we present the effectiveness of LiPCoT in classifying Parkinson's disease (PD) using an EEG dataset from 46 participants. In particular, we utilize LiPCoT to encode EEG data into a small vocabulary of tokens and then use BERT for self-supervised learning and the downstream task of PD classification. We benchmark our approach against several state-of-the-art CNN-based deep learning architectures for PD detection. Our results reveal that BERT models utilizing self-supervised learning outperformed the best-performing existing method by 7.1\% in precision, 2.3\% in recall, 5.5\% in accuracy, 4\% in AUC,  and 5\% in F1-score highlighting the potential for self-supervised learning even on small datasets. Our work will inform future foundational models for time series, particularly for self-supervised learning.
\end{abstract}

\section{Introduction}
Time series data, representing sequences of values over time, plays a vital role in diverse fields like finance, healthcare, weather, and sensor networks. However, analyzing and extracting insights from such data often requires specialized techniques. Traditional time series analysis methods heavily rely on domain-specific knowledge and feature engineering. Recent works explored recurrent neural networks (RNN) and convolution neural networks (CNN) for time series tasks, achieving promising results. However, these require significant computational resources, can struggle with capturing long-term dependencies, and aren't inherently suitable for self-supervised learning. On the other hand, transformer-based language models have recently shown outstanding performance in capturing long-term dependency, self-supervised learning, and computational efficiency. Thus, there is a need to integrate language models for time series analysis via self-supervised learning.

Self-supervised representation can offer unique benefits over supervised learning. First, supervised learning needs annotated data and is limited by the labeled data size and the quality of the labeling. Second, supervised models force a narrow learning of features for a single downstream task whereas self-supervised features can achieve better generalization for many downstream applications. However, there are some unique challenges in self-supervised learning for the time series domain compared to the natural language processing (NLP) of texts and similar transformer-based image models. This is mainly due to the fundamental difference in the nature of time series data, which are continuous-valued sequences, and text/images, which take discrete values from a finite vocabulary. Therefore, unlike NLP applications where word or sub-word tokens are used, there is no lexicon of discrete time series units, making it challenging to apply predictive losses in self-supervised learning.

This work proposes Linear Predictive Coding based Tokenizer for time series (LiPCoT), a novel tokenizer specifically designed to tokenize time series data for enabling self-supervised learning via language models. In particular, LiPCoT transforms time series data into a sequence of tokens, allowing existing language models like BERT to be leveraged for self-supervised training leading to downstream tasks like anomaly detection, forecasting, and classification. 

Instead of using CNN encoders which utilize temporal features, LiPCoT considers time series as a realization from an underlying stationary stochastic random process and creates a latent space of time series data using the parameters of the underlying random processes. This provides a stochastic representation of time series data from which discrete tokens are constructed. By utilizing the stochastic representation of time series, LiPCoT offers some unique benefits over other methods. For example, LiPCoT does not depend on the sampling frequency or length of the time series data which are crucial for other methods that utilize CNN encoders. 

In this paper, we present a proof-of-concept study where we propose LiPCoT and demonstrate the efficacy of LiPCoT in classifying Parkinson's disease (PD) using EEG data from 46 participants. We utilize LiPCoT for tokenizing EEG signals which are then leveraged by BERT for self-supervised learning and subsequent PD classification. We benchmark our approach against four state-of-the-art deep learning architectures, and our findings show that BERT models utilizing self-supervised learning on LiPCoT tokens outperform existing methods across all evaluated metrics for PD classification. 

The rest of the paper is organized as follows. Section \ref{sec:priorworks} discusses prior time series tokenization approaches in the literature. Section \ref{sec:lipcot} provides a detailed theory and methodology of LiPCoT. Section \ref{sec:exp} details our experiments for evaluating LiPCoT performance. The outcomes of our results are given in Section \ref{sec:results} and the ablation study is provided in Section \ref{sec:ablation}. Finally, Section\ref{sec:limitation}  discusses the limitations and future direction of our work, and Section  \ref{sec:conclude} concludes the paper.

\section{Related Works}\label{sec:priorworks}

So far, there have been limited attempts to convert time series data into discrete tokens using a discrete codebook. One of the fundamental approaches to this end is the quantization of time series data which converts a sequence of continuous numerical data into discrete representations. There are mainly two widely known approaches for this. The first one is Vector Quantized Variational AutoEncoder (VQ-VAE) which utilizes traditional encoder-decoder VAE architecture based on CNN to learn a discrete codebook via vector quantization technique. This approach has been utilized for time series encoding in various time series analyses and architectures including TOTEM, DeWave, Auto-TTE, UniAudio, and VioLA \cite{totem,dewave,autotte,uniaudio}. 

Another approach is to utilize a CNN encoder for extracting features from time series data which are then fed to a transformer-based architecture for generating a latent space via masked prediction. Finally, clustering techniques like k-means are utilized for creating a discrete codebook for time series data. This approach has been utilized in many transformer-based time series architectures such as HuBERT and Wav2Vec \cite{hubert,wav2vec}. 

Both of these approaches utilize CNN to reduce the length of the input time series sequences which are then fed to VAE or transformer-based encoder to generate a discrete codebook. Yet another approach to quantizing time series is by utilizing the frequency domain information. For example, FreqTST tokenizes time series data by first performing a Fourier transform to obtain the frequency spectrum and converts time series into discrete frequency units with weights \cite{freqtst}. 

Finally, a recent work proposed discrete wavelet transform for time series segmentation and dynamic time warping coupled with k-means for vocabulary creation  \cite{ecgbert}. While there are some variations in these approaches, none of them utilize the stochastic nature of time series data during the encoding process.  

\section{LiPCoT: Tokenization of time series data}\label{sec:lipcot}
\subsection{Objective}
The primary objective of our approach is to provide a tokenization method of time series data such that it is readily compatible with the existing NLP language models for self-supervised learning. To this end, we propose a novel tokenization approach that can take time series data and convert them into discrete tokens that can be leveraged by language models such as BERT via pre-training and fine-tuning. 
\subsection{Overview}
Fundamentally, we assume that the time series data is piece-wise stationary and divide time series data into segments. Then, assuming each segment as a realization from a stationary stochastic random process, we estimate the underlying random process and create a latent space of time series data using the parameters of the random processes. This gives us a representation of time series data from which we create tokens by clustering the aforementioned latent space. Finally, we feed these tokens to a language model for pre-training and fine-tuning tasks.
\begin{figure}\label{Fig1}
	\centering
	\includegraphics[width=13.5cm,height=13cm,clip,keepaspectratio]{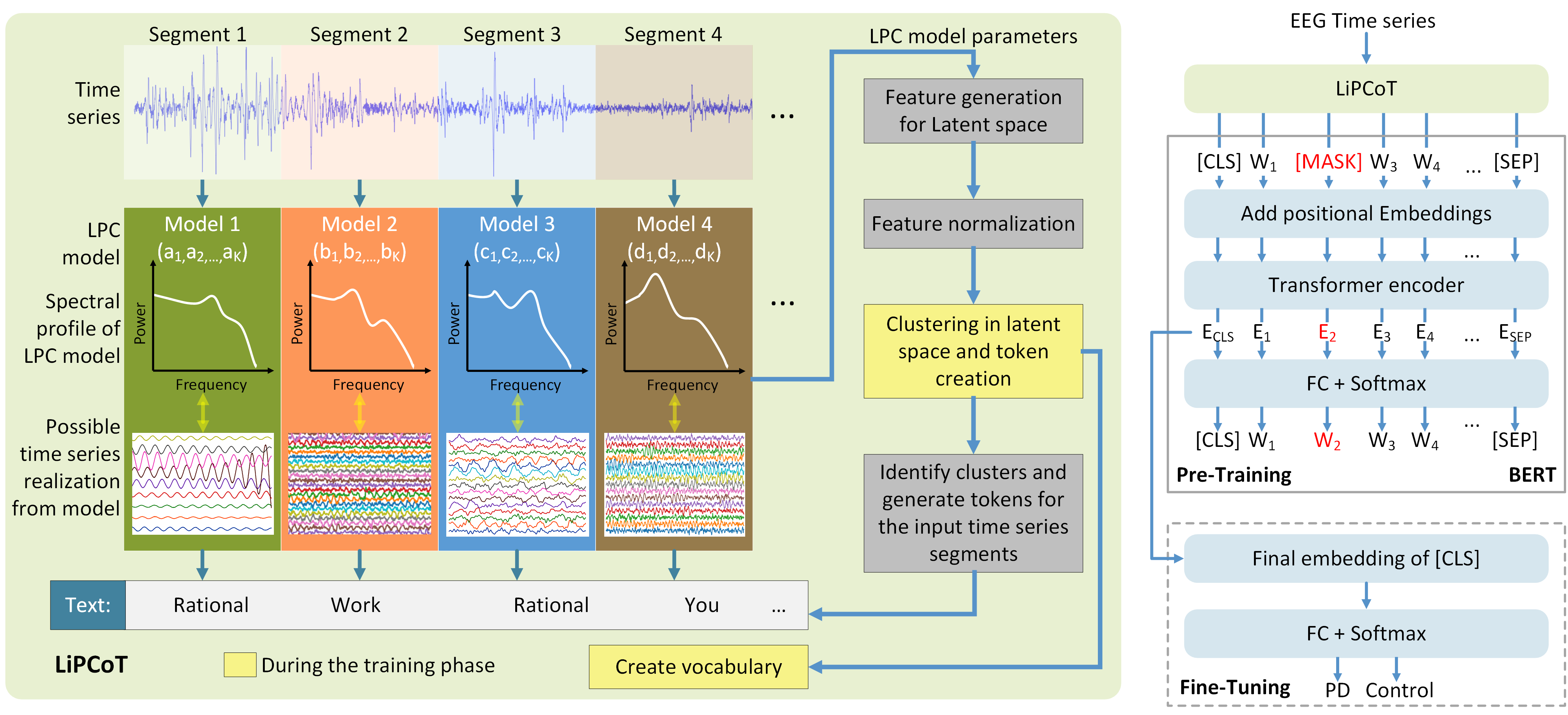}
	\caption{Overview of LiPCoT and its application for PD classification via BERT.}
\end{figure} 
\subsection{Stochastic modeling of time series}
\subsubsection{Linear Predictive Coding}
Our proposed approach utilizes Linear Predictive Coding (LPC), a widely used technique in signal processing and speech analysis for estimating the stochastic random process of time series. 
One of the key advantages of LPC is its ability to compactly represent the spectral characteristics of a signal using a small number of coefficients. This compression of signals via stochastic modeling makes it a powerful tool for predicting the behavior of distinguishing time series \cite{RN3}. LPC is one of the dominant analyzing techniques in speech processing, enhancement, and coding \cite{RN1,RN24,RN27,RN34,RN35}. It has also been used in EEG coding \cite{RN19}, economics \cite{RN28}, control theory \cite{RN11}, filtering \cite{RN15}, and a host of other applications. 

At its core, LPC fits an autoregressive (AR) model to a time series. Specifically, suppose one has the time series sequence:
\begin{equation}
	x_0,x_1,...,x_{N-1}
	\label{e1}
\end{equation}
with subscripts representing the sample indices. The $L^{th}$ order LPC model of such a Wide Sense Stationary (WSS) time series provides an autoregressive (AR) approximation of the data. In particular, with $ \eta_n $ as a white driving sequence that comes from a white WSS process $\eta$ with variance $\sigma^2$, LPC approximates $ x_n $ as the output of a predictor that uses a linear combination of its past samples.
\begin{equation}
	\hat{x}_n=-\sum_{k=1}^{L} a_{k} x_{n-k}+\eta_n
	\label{eq:LPC}
\end{equation}
The coefficients $a=(a_1,a_2,\dots,a_{L})$ are the parameters of LPC model known as LPC coefficients and are calculated by minimizing the prediction error, $\mathbb{E}[(x_n-\hat{x}_n)^2]$.

Using the z-domain, one can see that LPC approximates $x_n$ using $ \eta_k $ and a predictor with transfer function
\begin{equation}
	H(z)=\frac{1}{1+\sum_{k=1}^{L}a_{k} z^{-k}}
	\label{eq:Hz}
\end{equation}
 where $z^{-1}$ denotes unit delay shift operation. Note that, many algorithms exist for the calculation of LPC coefficients $a_{k}$. Two alternative characterizations are possible for describing the predictor transfer function $H(z)$. The first uses the LPC coefficients, $a=(a_1,a_2,\dots,a_L)$ as shown in (\ref{eq:Hz}). The second uses $p=(p_1,p_2,\dots,p_L)$, the poles of the transfer function:
\begin{equation}
	H(z)=\prod_{k=1}^{L}\frac{1}{(1-p_k z^{-1})}.
	\label{e3}
\end{equation}
It can be seen that the poles are roots of the polynomial $1+\sum_{k=1}^{L}a_{k} z^{-k}=0$. 

One of the advantages of LPC is its ability to efficiently approximate the spectral density from time series. Specifically, one can obtain the estimate of spectral density by.
\begin{equation}
	P(f)=\sigma^2|H(e^{\nicefrac{j 2 \pi f}{F_s}})|^2.
	\label{eq:PSD}
\end{equation}
where $F_s$ is the sampling frequency. It is well known, \cite{LPC_stability}, that as long as the $ L $-th order autocorrelation matrix of $ x_n $ is positive definite, $ H(z) $ has all poles in the open unit unit disk $|p_k| \leq 1$. For real-valued time series $x_n$, the poles $p_k$ come as conjugate pairs and their phase provides an estimate of the dominant frequency component characterizing the time series. Specifically, if the time series has a dominant spectral component (peak in the power spectrum) at frequency $f_{oc}$ Hz, then the LPC poles will be of the form, 
\begin{equation}
	p_k=A e^{\pm \nicefrac{j 2 \pi f_{oc}}{F_s} }
	\label{p_k}.
\end{equation}

\subsubsection{Frequency-warped Linear Predictive Coding}
The spectral density captured by LPC uniformly covers all frequencies within the sampling frequency range of a time series. However, in practice, useful information often is not uniformly distributed across all frequencies but rather localized in higher or lower frequencies. A good example of this phenomenon is the brain activity signals where we observe $\nicefrac{1}{f}$ characteristics in the frequency domain where most informative activities occur in low-frequency ranges. 
 
Frequency-warped linear predictive coding is a variation of LPC that estimates spectral powers in a non-uniform resolution \cite{harma2000frequency,harma2000evaluation,roth2003frequency}. It includes a warping coefficient $\lambda \in [-1,1]$ which enables it to provide higher resolution at low-frequency powers for $\lambda>0$ or at high-frequency powers for $\lambda<0$. At  $\lambda=0$ it becomes the traditional LPC with a uniform resolution for all frequencies (see details in Appendix \ref{apdx_2}).

\subsection{Latent space for time series}\label{subsec: latent space}
Here, We describe how univariate time series data can be represented in a latent space based on stochastic modeling of the time series via Frequency-warped LPC. In particular, we use Burg's method for calculating the LPC coefficients\cite{roth2003frequency}. We augmented the algorithm proposed in \cite{roth2003frequency} in two ways (Algorithm \ref{algo1}). First, we added the estimation of the power of prediction error $\sigma^2$ and implemented a fast version of the traditional Burg's algorithm proposed in \cite{vos2013fast}. Note that the conventional implementation of Burg's method has the complexity of $\mathcal{O}(NL)$ which can be reduced to $\mathcal{O}(NlogN+L^2)$ using Fast Fourier Transform\cite{vos2013fast}.

\begin{algorithm}\label{algo1}
	\caption{Proposed Burg's method for Frequency-warped LPC}
	\SetAlgoLined
	\KwResult{$a$, $\sigma^2$}
	\KwData{timeseries $x: [x_0,x_1,\dots x_{N-1}]$, order $L$, Warping coefficient $\lambda$}
	\textbf{Initialization:} 	$a\leftarrow 1$, 
		 $b \leftarrow x$,	 $f \leftarrow x$, $\sigma^2\leftarrow \frac{xx^T}{N} $\\
	\For{$i=1,2,\dots,L$}{
		$\hat{b}\leftarrow 0_{1\times (N-i)}$\\
		$\hat{b}_0\leftarrow b_0-\lambda b_1$\\
		\For{$j=1,2,\dots,N-i$}{
			$\hat{b}_j\leftarrow b_j-\lambda (b_{j+1}-\hat{b}_{j-1})$
		}
		$\hat{f}\leftarrow [f_1,f_2,\dots,f_{N-i}]$\\
		$k\leftarrow \frac{-2\hat{f} \hat{b}^H}{\hat{f}\hat{f}^H+\hat{b}\hat{b}^H}$\\
		$f\leftarrow \hat{f}+k\hat{b}$\\
		$b\leftarrow \hat{b}+k^*\hat{f}$\\
		$\sigma^2\leftarrow (1-|k|^2)\sigma^2$\\		

		$a\leftarrow \begin{bmatrix}
			a \\
			0 \\
		\end{bmatrix}+k %r_i
		J \begin{bmatrix}
		a^* \\
		0 \\
		\end{bmatrix}$, where $J=\begin{bmatrix}
		0 		&  \dots 	& 0 	& 1 \\
		\vdots 	&  \iddots 	& 1 	& 0 \\
		0 		&  \iddots 	&\iddots& \vdots \\
		1 		&  0 		& \dots & 0 \\
		\end{bmatrix}_{(i+1) \times (i+1)}$
	}
	$a \leftarrow [a_1,a_2,\dots,a_{L}]$, where $a=[1,a_1,\dots,a_{L}]$
\end{algorithm}

Next, we extract features from the frequency-warped LPC model to construct our Latent space. One desired property we seek is to fully recover the LPC models from the feature space. There are a few ways this can be achieved:

\subsubsection{LPC coefficients}\label{lpc_coeff}
We can use the weighted LPC coefficients where the weights are $w_1,w_2,\dots, w_L$ to create a $L$-dimensional latent space. However, this formulation is agnostic to total signal power. Hence, We propose an extended $L+1$-dimensional space by adding the power of prediction error. In particular, a feature vector for $x$ is,
\begin{equation}
	F_1(a):=(w_1 a_1,w_2 a_2,\dots,w_L a_L,\text{log}\sigma^2).
	\label{F1}
\end{equation}
The distance metric between two LPC models $a$ and $a'$ is,
\begin{equation}
	d_{\text{COEF}}(x,x'):=\sqrt{(\text{log}\sigma^2-\text{log}\sigma'^2)^2+\sum_{i=1}^{L} w^2_i (a_i-a'_i)^2}.
\end{equation} 
The rationale for using weights is that not all $a_i$ has the same impact on the AR model. However, defining a good set of weights is a hard problem \cite{martin2000metric}. In this study, we assume $w_i=1; \forall i$. However, we note that the weights can be optimized using any appropriate cost function in a self-supervised architecture. This will be investigated in future studies.

\subsubsection{Cepstrum coefficients}
The cepstrum of a stochastic random process is defined by the inverse Fourier transform of the log of the power spectrum of the process \cite{boets2005clustering}. For a stochastic model such as LPC, the cepstrum of the output process $(c_0,c_1,\dots,c_n,\dots)$ can be calculated from model parameters\cite{boets2005clustering,kalpakis2001distance}. We create our latent space by taking the first $M$ weighted cepstrum coefficients (Appendix \ref{apdx_3}),
 \begin{equation}
 	F_2(a):=(c_0,c_1,\sqrt{2}c_2,\dots,\sqrt{M}c_M)
 	\label{F2}
 \end{equation}
\subsubsection{Dominant spectral components}
With the limitations of the aforementioned approaches, we propose yet another latent space with $2L+1$ dimensions termed dominant spectral components. The salient point of the latent space is to create a space based on the dominant spectral components as determined by the poles $p_k$ in (\ref{p_k}) where the dominant frequency is $f_0$. However, the poles $p_k$ have no particular order. Therefore, first, we order the poles based on the frequencies of the dominant spectral components and then construct the latent space with the angle and radius of the poles as well as the prediction error $\sigma^2$ from Algorithm \ref{algo1}, 
\begin{equation}
	F_3(a):=\big(u_1, u_2,\dots,u_L, v_1,v_2,\dots,v_L,\text{log}\sigma^2\big)
	\label{F_my}
\end{equation}
where,
\begin{equation}
	u_k=\frac{F_s}{2\pi} \angle p_k,
\end{equation}
\begin{equation}
v_k=-2\text{log}(1-|p_k|)
\end{equation}
with the ordering scheme,
\begin{equation}
	|u_i|\leq |u_j|\qquad \text{if } i<j.
\end{equation}

The first $L$ components in this latent space are the dominant frequencies and the next $L$ elements are analogous to the power in the respective dominant frequencies (see Appendix \ref{apdx_1}). Finally, the last component is the power of the white noise process. This definition of the latent space in (\ref{F_my}) is inspired by how we naturally interpret power spectrum plots which makes the latent space more interpretative.

Few things to note here. First, for zero mean time series data, all poles are complex and come in conjugate pairs. Hence, there are duplicates of values within $v_i$ in (\ref{F_my}) due to the conjugate pair of poles which can be discarded for an equivalent but smaller $L+1$ dimensional latent space for low-pass filtered data. Second, the latent space is agnostic of the sampling frequency $F_s$ and signal length, which is a desired quality for a general time series tokenization model.

\subsection{Tokenization of time series}
\subsubsection{Data Segmentation}
As time series data can have a variable length, first we divide the time series into segments of a fixed window and we fit a LPC model for each segment. These segments can be overlapping or non-overlapping.
        
\subsubsection{Token generation}
During training, we calculate LPC model $a$ for each time series segment $x$ using Algorithm \ref{algo1} and generate a latent space based on (\ref{F1}),(\ref{F2}) or (\ref{F_my}) such that each LPC model $a$ is projected into the latent space $F_i(a)$. Next, we cluster the space for the quantization of the LPC models such that $F_i(a)\in \hat{F}_k; \forall a$ for some $k\in\{1,2,\dots,K\}$ and assign a unique token (and a particular word) to each cluster $\hat{F}_k$ resulting in a vocabulary $\{ \hat{F}_1, \hat{F}_2,\dots,\hat{F}_K\}$. For this, first, we normalize each dimension of the latent space during training and utilize an unsupervised k-means clustering algorithm. 

\subsubsection{Encoding}
The encoding step is similar to the token generation where for each time series segment we calculate the LPC model and obtain a representation $F_i(a)$ in the latent space. However, the corresponding cluster is estimated by using the previously trained k-means clusters. This provides a unique token for the time series segment. Similarly, tokens are generated for all segments of the time series. 

\subsubsection{Decoding}
As we use a stochastic model of the time series for tokenization, recovering the exact time series is not possible. However, we can obtain a realization of the stochastic source of the time series. To estimate a time series segment from a token or word, we use white noise as the primary source which is then filtered appropriately to match the stochastic nature of the desired time series segment. First, we find the corresponding cluster center $\hat{F}_k$ from the given token which gives an estimation of the LPC model $\hat{a}$ and the noise power $\hat{\sigma}^2$ for the time series segment. Now, we construct the estimated transfer function $\hat{H}(z)$ and a WSS process $\eta$ with variance $\hat{\sigma}^2$. Finally, the time series estimation $\hat{x}$ is obtained by filtering a   WSS realization sequence $\eta_n$ with $\hat{H}(z)$.

\subsection{Integration with Language Models}
As LiPCoT converts time series into tokens with a corresponding vocabulary, integrating into an NLP-based language model is relatively simple and can be done in two ways. First, we can pre-train a language model such as BERT \cite{devlin2019bert} from scratch with the given vocabulary and training data. This can lead to a language model capable of analyzing time series data. Another way of integrating LiPCoT is to take a pre-trained language model and add the new tokens from LiPCoT to its existing vocabulary. In this case, the embedding space has to be resized and the model needs to get further pre-training to generate embedding for the newly added timeseries related words. In this work, we focus on the first method.

\subsubsection{Self-supervised learning}
During the pre-training stage, we utilized the traditional Masked Language Modeling (MLM) to train our BERT model via self-supervised learning (Figure \ref{Fig1}). In particular, MLM was implemented by masking 15\% of the words randomly where 80\% of the words with "[MASK]" token, 10\% with some other random words, and the rest 10\% were unchanged.

\subsubsection{Fine-tuning task: binary classification}
In this study, we focused on binary classification of Parkinson's disease (PD) from EEG data as a downstream task. To achieve this, we fine-tuned the model by first obtaining the final hidden embedding layer for the "[CLS]" class token from the pre-trained BERT model and then adding a fully-connected layer to this with a sigmoid function (Figure \ref{Fig1}). Finally, we normalized the outputs of the sigmoid function to obtain the probability of binary classification. 

\section{Experiments}\label{sec:exp}
\subsection{Dataset}
For our experiments, we used an EEG dataset of 54 participants from a study at the University of New Mexico (UNM; Albuquerque, New Mexico) where 27 had PD and the rest of the participants were healthy which was previously described in \cite{anjum2020linear}. Upon manual inspection, we utilized EEG data from 46 participants (22 PD and 24 healthy subjects). EEG data were recorded with a sampling rate of 500 Hz on a 64-channel Brain Vision system. PD patients were in OFF medication state.

\subsection{Preprocessing of data}
In this work, We utilize EEG data from 59 channels out of 63 based on average channel data quality. The data from each channel were high-pass filtered at 1 Hz to remove noise. No other pre-processing was implemented. Only the first one minute of EEG data from each participant were utilized which corresponds to eyes closed resting state EEG.

The multi-channel data ($5n$ seconds) for each subject ($\mathbb{R}^{59\times 5n F_s}$) were converted into 5-second segments ($\mathbb{R}^{n\times 59\times 5F_s}$). For each 5-second segment, LiPCoT was utilized to convert the time series data for each channel into one token resulting in a single sequence of $59$ tokens where the location of each channel in the sequence was fixed. This approach of constructing sequences ($\mathbb{N}^{n\times 59 \times 1}$) was performed to encode the spatial embedding of EEG channels into the positional embedding of each sequence of tokens. 

\subsection{Experiment setup}
First, we randomly shuffled data at the subject level and split the dataset into training (60\%), validation (20\%), and test (20\%) datasets. We utilized the training data without classification labels for self-supervised training via BERT using MLP. The validation data without labels were used for evaluating the model's performance against overfitting and the best-performing model on the validation set was selected. For the PD classification task, we used the validation data with labels for training. To evaluate the model's performance, we utilized the training data with classification labels and selected the best-performing model. To measure the classification performance, we utilized five metrics: precision, recall, accuracy, F1-score, and AUC. The classification performance was evaluated on the test dataset. 
 
\subsection{Comparison with state-of-the-art supervised learning }
We investigated whether fine-tuning a self-supervised BERT model through LiPCoT tokens can outperform traditional state-of-the-art CNN-based models that are trained on raw time series data via supervised learning for the downstream PD classification task. To achieve this, we utilized four CNN architectures that have been shown to perform well in PD classification using EEG data: 13-layer Deep CNN \cite{deepcnn}, ShallowConvNet \cite{convnet}, DeepConvNet \cite{convnet} and EEGNet \cite{eegnet}. We chose these methods as they were shown to be very effective neural network architectures tailored for EEG-based PD classification in the literature. Unlike BERT which was trained on tokenized data, these CNN-based models were trained on continuous-valued time series data without any tokenization. The input to these state-of-the-art models were 5-second time series data segments from $59$ channels ($\mathbb{R}^{n\times 59\times 5F_s}$) with corresponding labels. We used the validation data with labels to train these models.

\subsection{Model Parameters}
For training LiPCoT, we utilized the training dataset without labels. LPC models were 16\textsuperscript{th} order ($L=16$). The total vocabulary length was set to 64. Additionally, the warping coefficient $(\lambda)$ was 0.2, and the Latent space was generated using LPC coefficients. We initialized the BERT model with 6 hidden layers each with 256 neurons. We utilized relative position for increased robustness and to eliminate the limitation of token length. The total parameter size was 11,356,485 (Appendix \ref{apdx_4}). During self-supervised learning, BERT was trained with 256 epochs and a batch size of 2. We utilized the Bayesian optimization method to determine the optimal learning rate and batch size for the downstream classification task. The optimal batch size was 4 and the learning rate was $1.9\times 10^{-5}$. The training was conducted for 64 epochs. 

\section{Results}\label{sec:results}
\begin{figure}[htbp]
\label{AF0}
\centering
\includegraphics[width=14cm,height=10.5cm,clip,keepaspectratio]{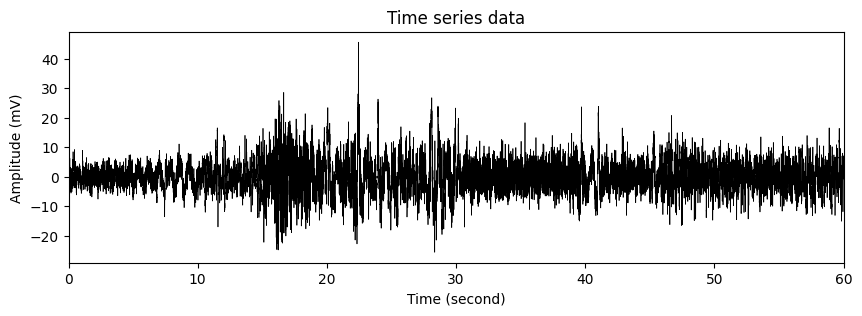}
\includegraphics[width=14cm,height=10.5cm,clip,keepaspectratio]{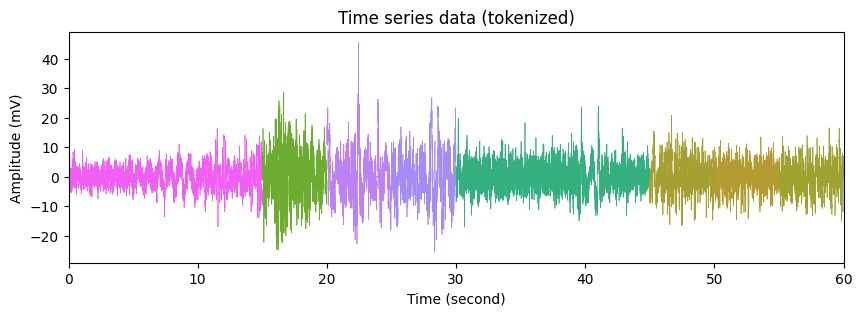}
\caption{Tokenization of time series data via LiPCoT: One-minute data from the validation set from a single EEG channel before (top) and after (bottom) tokenization. Each color represents a unique token. LPC coefficients were utilized for latent space construction with order $L=16$, warping coefficient $\lambda=0.2$.}
 \end{figure}

\begin{figure}[htbp]
\label{AF1}
\centering
\includegraphics[width=14cm,height=10.5cm,clip,keepaspectratio]{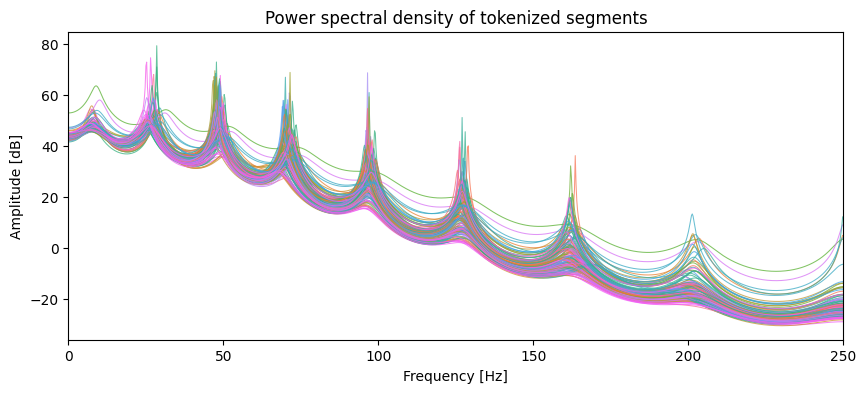}
\caption{Spectral density of tokenized data segments: Power spectral density of 5-second data segments in a single EEG channel from the validation set colored by their respective LiPCoT tokens. LiPCoT with LPC coefficients, order $L=16$, warping coefficient $\lambda=0.2$. }
 \end{figure}

\begin{figure}[htbp]
\label{AF4}
\centering
\includegraphics[width=13.5cm,height=10.5cm,clip,keepaspectratio]{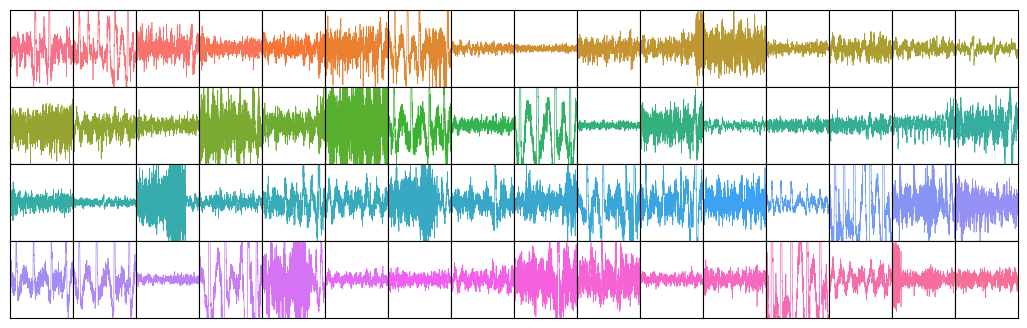}
\caption{Tokenized data segments: Representative data segments colored by their respective LiPCoT tokens. Each plot shows a single 5-second time series segment. Data from a single EEG channel in the validation set. LiPCoT with LPC coefficients, order $L=16$, warping coefficient $\lambda=0.2$.}
 \end{figure}

Figure \ref{AF0} depicts an example of time series tokenization via LiPCoT. The spectral density of the tokenized data segments showed variations in the spectral profile of the tokens (Figure \ref{AF1}). This resulted in the effective capturing of temporal changes in time series data by LiPCoT tokens (Figure \ref{AF4}).

\begin{table}[htbp]
	\caption{Performance comparison}
	\label{perf}
	\centering
	\begin{tabular}{lllllll}
		\toprule
		 Architecture  & Input data & Precision  & Recall  & F1-score & AUC & Accuracy \\
		\toprule
        DeepCNN\cite{deepcnn}  & TS & 48.8			& 47.7			&0.48  & 0.54 & 51.1\\
         ShallowConvNet\cite{convnet} & TS & 75.7 & 56.8 & 0.65 & 0.77 & 70.6\\
         DeepConvNet\cite{convnet} & TS & 68.9 & 70.4 & 0.70 & 0.79 & 70.6\\
         EEGNet-8,2\cite{eegnet} & TS & 69.6 & 72.7 & 0.71 & 0.78 & 71.7\\
		 BERT (Ours) & LiPCoT tokens & \textbf{76.7}		& \textbf{75}	& \textbf{0.76} &\textbf{0.82}	&\textbf{77.2}\\
        
		\bottomrule
		\multicolumn{7}{l}{\footnotesize TS = time series; Best performance in bold.}\\
	\end{tabular}
\end{table}

Our results show that BERT models with self-supervised learning outperformed the state-of-the-art architectures in all metrics. Among the four architectures compared in this study, EEGNet provided the best overall performance. Our BERT model with self-supervised learning  outperformed EEGNet by 2.3\% in recall, 7.1\% in precision, 4\% in AUC, 5\% in F1-score, and 5.5\% in accuracy (Table \ref{perf}).

\section{Ablation study}\label{sec:ablation}

\subsection{Optimal tokenization approach for LiPCoT}
In Section \ref{subsec: latent space}, we have provided three approaches to construct a latent space via LPC for LiPCoT tokenization. We investigated the effectiveness of these approaches for extracting features from time series. In other words, we aimed to find the best approach among the three for time series tokenization suitable for self-supervised learning and the PD classification task. For this, We utilized these approaches for the tokenization step before self-supervised training of BERT and compared the performance in PD classification. Our results show that among the aforementioned methods for constructing latent space for LiPCoT, LPC coefficients (detailed in Section \ref{lpc_coeff}) were the most effective latent features for LiPCoT in the downstream PD classification task. In particular, they outperformed the other methods by up to 10.9\% in accuracy, 11\% in AUC, and 19\% in F1-score (Table \ref{ablation}). The cepstrum coefficients provided similar performance to LPC coefficients with higher precision but a lower recall rate. Dominant spectral coefficients showed significantly lower performance than the rest of the methods indicating that the latent space for this method is not inherently Euclidean and standard k-means is not a suitable method for clustering the space. 

\begin{table}
	\caption{Ablation study}
	\label{ablation}
	\centering
	\begin{tabular}{lllllll}
		\toprule
		\multicolumn{2}{c}{Model details} & \multicolumn{5}{c}{Classification performance}\\
		\cmidrule(r){1-2} \cmidrule(r){3-7}
		LiPCoT method  & Self-supervised  & Precision  & Recall  & F1-score & AUC & Accuracy \\
		\toprule
           DSC				& No		& 60   & 47.7 & 0.53 & 0.61	&59.8\\
           Cepstrum coeff.	& No		& 78.8 & 59.1 & 0.67 & 0.76	&72.8\\  
           LPC coeff.		& No		& 65.9 & 65.9 & 0.66 & 0.71	&67.4\\  
		\midrule
           DSC				& Yes		& 72.4 & 47.7 & 0.57 & 0.71	&66.3\\
           Cepstrum coeff.	& Yes		& \textbf{82.3} & 63.6 & 0.72& 0.79	&76.1\\	
		   LPC coeff.		& Yes		& 76.7		& \textbf{75}	& \textbf{0.76} &\textbf{0.82}	&\textbf{77.2}\\	
		\bottomrule
		\multicolumn{7}{l}{\footnotesize DSC= Dominant spectral components}\\
	\end{tabular}
\end{table}
 
 \subsection{Effectiveness of self-supervised learning on tokenized data}
 We also investigated whether self-supervised learning provides any significant advantage for the supervised classification task on the LiPCoT tokenized time series data. For this, we utilized an untrained BERT model initialized with random weights and biases as a baseline for the pre-trained BERT model trained via MLM self-supervised learning. Both were utilized in the fine-tuning stage for the PD classification task and their performances were compared. This paved the way to measure how much information a self-supervised BERT model can add to a downstream supervised classification model when deployed on time series data tokenized by LiPCoT.

 We found that self-supervised learning via BERT significantly boosted the performance in supervised classification tasks for PD detection. In particular, when compared to BERT models initialized with random seeds, models with self-supervised training showed performance enhancement up to 9.8\% in accuracy, 10.8\% in precision, 9.1\% in recall, 10\% in F1-score and 11\% in AUC (Table \ref{ablation}). The highest performance boost resulted from the LPC coefficients. Recall that the supervised training was conducted on a validation dataset which was only 3 times smaller than the training dataset of self-supervised learning. Therefore, these results demonstrate that even on a small scale, LiPCoT has the potential to effectively tokenize time series data that can boost performance for supervised classification via self-supervised learning in unlabeled data.

\subsection{Fourier Transform vs. LPC}
Stochastic modeling via LPC can provide an envelope of the power spectrum of the time series (Appendix \ref{apdx_1}). One advantage of the LPC-based power spectrum over traditional Discrete Fourier transform (DFT) is its dynamic frequency resolution. Unlike DFT, where frequency resolution is uniform across 0 Hz to $\nicefrac{F_s}{2}$ Hz and depends on the number of data points used, LPC-based power spectrum allows evaluation at any frequency without being affected by the sequence length resulting in more accurate detection of major oscillations compared to DFT. Furthermore, frequency-warped LPC enables us to further emphasize higher or lower frequency (depending on the warping coefficient $\lambda$; Appendix \ref{apdx_2}). Another advantage of stochastic modeling is its ability to compress a time series into a few LPC parameters by estimating the underlying random process. This means that it is invariant of time shifts and can be made agnostic to scaling of time series when necessary.

\subsection{LiPCoT vs. CNN}
While many of the existing architecture utilizes CNN for generating features from time series data, LiPCoT uses LPC which encodes data via stochastic modeling. Hence, the input data for LiPCoT can be of variable lengths with different sampling frequencies which is not possible for a CNN-based feature generation. This is especially important for creating foundation models with large datasets. Furthermore, the latent space of LiPCoT is shift invariant and interpretative. Finally, the tokenization of LiPCoT is computationally efficient making it suitable for large-scale deployments.

\section{Limitations and Future Directions}\label{sec:limitation}
One major limitation of our work is the lack of a bigger and more diverse dataset which could highlight the implications of our approach for a more generalized time series classification. However, due to the limited computational power and scarcity of similar datasets, in this proof-of-concept work, we focused on the feasibility of our approach and measured the key advantages of our architecture in the downstream task performance. To this end, even if our dataset was limited, we were able to observe a significant benefit of self-supervised learning of time series data enabled by LiPCoT tokenization and superior performance in the downstream task of PD classification compared to traditional approaches. Our results indicate that with a sufficiently large dataset for the pre-training, the performance can go even higher. 

Another limitation of our proposed LiPCoT tokenizer is its inability to fully recover the original time series after tokenization. The tokens in LiPCoT capture the underlying stochastic random process. This can possibly limit its effectiveness for short-term forecasting tasks. However, such stochastic representation of time series data can be beneficial for long-term forecasting. 

It should be noted that the process of forecasting and generation of time series through LiPCoT is very similar to the diffusion model as both of them generate outputs from white noise. Conceptually, one can generate many 'candidate' time series predictions using the LPC models embedded in the latent space of LiPCoT and choose the best candidate via optimization of prediction error. These aspects will be further investigated in a future study.

\section{Conclusion}\label{sec:conclude}
In this work, we propose LiPCoT, a tokenizer for time series data that converts time series signals into discrete tokens via stochastic modeling. We measured LiPCoT's performance by utilizing BERT for self-supervised learning and the downstream PD classification task on tokenized time series data. We compared the performance of our downstream task with four state-of-the-art CNN-based architectures. We used a relatively small dataset for our experiments compared to the typical size required for self-supervised training with transformer-based models like BERT.  Despite this, our results showed that by utilizing the data tokenized by LiPCoT, self-supervised learning via BERT resulted in 10.8\% in precision, 9.1\% in recall, 9.8\% in accuracy, 10\% in F1-score, and 11\% in AUC improvement in supervised classification task of PD detection. Our approach outperformed the state-of-the-art models for PD classification that utilize time series data without any tokenization by 7.1\% in precision, 2.3\% in recall, 5.5\% in accuracy, 4\% in AUC, and 5\% in F1-score.
 
\subsection*{Data and Code Availability}
The original EEG dataset can be found at \href{http://predict.cs.unm.edu/downloads}{http://predict.cs.unm.edu/downloads.php}. The pre-processed EEG dataset in .csv formats can be found via this Dropbox \href{https://www.dropbox.com/scl/fi/xinqn33vof0bnb9rlvmdh/raw.zip?rlkey=jb4dyumh7v82vbj36wsb53x13&dl=0}{link} and the Python codes are in \href{https://github.com/MDFahimAnjum/LiPCoT}{https://github.com/MDFahimAnjum/LiPCoT}.

%\subsubsection*{Acknowledgments}
%Use unnumbered third level headings for the acknowledgments. All acknowledgments go at the end of the paper. Do not include acknowledgments in the anonymized submission, only in the final paper.

\bibliographystyle{plain} % Use plain style for numbered citations
\setlength{\itemindent}{0pt} % Set the first line indent to 0
%\bibliography{bibliography.bib} % for local

%\begin{appendices}
%\pagebreak
%\newpage
\appendix
\section{Appendix}
\subsection{Power spectrum from LPC model}\label{apdx_1}
In this section, we discuss the relationship between power spectrum and LPC model. First, we investigate a second order LPC model where $L=2$ in (\ref{eq:LPC}), 
\begin{equation}
	\hat{x}_n=- a_{1} x_{n-1}- a_{2} x_{n-2}+\eta_n
\end{equation}
with two LPC coefficients $a=(a_1, a_2)$ and two poles $p=(p_1,p_2)$. For real-valued signals, poles are either
real or in complex conjugate pairs. Poles with negative imaginary part result from the mathematical symmetry of polynomials with real coefficients and represent the poles of the negative frequencies. Note that for real-valued signals, power spectrum and Fourier transform are symmetrical for positive and negative frequency. Thus, from (\ref{p_k}), we can express $p_1$ and $p_2$ as,
\begin{equation}
	p_1=A e^{ \nicefrac{j 2 \pi f_{oc}}{F_s} }
\end{equation} 
\begin{equation}
	p_2=A e^{ -\nicefrac{j 2 \pi f_{oc}}{F_s} }
\end{equation}
where $A\leq 1$ for a stable model. 

\begin{figure}[ht]\label{AF3}
	\centering
	\includegraphics[width=10.5cm,height=14cm,clip,keepaspectratio]{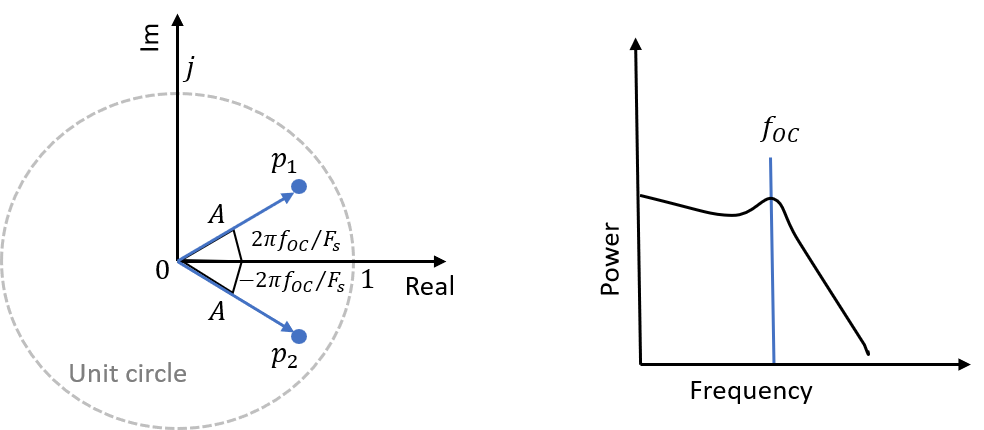}
	\caption{Illustration of poles from a second order LPC model in z-domain (left) and the corresponding power spectrum with one dominant frequency peak (right).}
\end{figure}

Now, from (\ref{eq:Hz}) and (\ref{e3}) we see the transfer function,
\begin{align}
	H(z) &= \frac{1}{1+ a_{1} z^{-1}+ a_{2} z^{-2}}\\
		 &= \frac{1}{(1-p_1 z^{-1})(1-p_2 z^{-1})}
\end{align}
and from (\ref{eq:PSD}) we obtain the log spectral power,
\begin{align}
	\text{log} P(f) &=\text{log} \sigma^2+\text{log} |H(e^{\nicefrac{j 2 \pi f}{F_s}})|^2\notag\\
		 &= \text{log} \sigma^2-  \text{log} \left| 1-p_1 e^{-\nicefrac{j 2 \pi f}{F_s}}\right|^2-  \text{log}\left|1-p_2 e^{-\nicefrac{j 2 \pi f}{F_s}} \right|^2\notag\\
		%&= \sigma^2 \left| \frac{e^{\nicefrac{j 4 \pi f}{F_s}}}{(e^{\nicefrac{j 2 \pi f}{F_s}}-p_1 )(e^{\nicefrac{j 2 \pi f}{F_s}}-p_2 )} \right|^2\\
		 &=\text{log} \sigma^2-  \text{log} \left| 1-A e^{\nicefrac{j 2 \pi (f_{oc}-f)}{F_s}}\right|^2-  \text{log} \left|1-A e^{-\nicefrac{j 2 \pi (f_{oc}+f)}{F_s}}\right|^2\notag\\
		 %&= \sigma^2 \left| \frac{e^{\nicefrac{j 4 \pi f}{F_s}}}{(e^{\nicefrac{j 2 \pi f}{F_s}}-A e^{ \nicefrac{j 2 \pi f_{oc}}{F_s} } )(e^{\nicefrac{j 2 \pi f}{F_s}}-A e^{ -\nicefrac{j 2 \pi f_{oc}}{F_s} } )} \right|^2
		 &=\text{log} \sigma^2-  \text{log} \left|1+A^2-2A\text{cos}(\nicefrac{2 \pi (f_{oc}-f)}{F_s})\right|\notag\\
		 &\quad- \text{log} \left|1+A^2-2A\text{cos}(\nicefrac{2 \pi (f_{oc}+f)}{F_s})\right| 
\end{align}
Therefore, the spectral density has a peak at $\pm f_{oc}$ Hz with power,
\begin{align}
	\text{log} P(f)|_{f=\pm f_{oc}} &=\text{log} \sigma^2-  2\text{log} \left|1-A\right|- \text{log} \left|1+A^2-2A\text{cos}(\nicefrac{4 \pi (f_{oc})}{F_s})\right|
\end{align}
Therefore, a contribution of each pole to the log power spectrum at $\pm f_{oc}$ Hz  is $-  2\text{log} \left|1-A\right|$.

Now, let us assume the poles are both real such that $p_1=A_1$ and $p_2=A_2$. Then, we have a peak at 0 Hz with power,
\begin{align}
	\text{log} P(f)|_{f=0} &=\text{log} \sigma^2 -2\text{log} \left|1-A_1\right|-  2\text{log} \left|1-A_2\right|
\end{align}
Therefore, the contribution of each pole to the log power spectrum is again $-  2\text{log} \left|1-A_i\right|$.

In general, for any LPC model with $L$ order we can write the poles as,
\begin{equation}
	p_i=A_i e^{ \nicefrac{j 2 \pi f_{i}}{F_s} }
\end{equation}
and the log of spectral power is,
\begin{align}
	\text{log} P(f) %&=\text{log} \sigma^2+\text{log} |H(e^{\nicefrac{j 2 \pi f}{F_s}})|^2\notag\\
	%&= \text{log} \sigma^2- \sum_{i=1}^{L} \text{log} \left| 1-p_i e^{-\nicefrac{j 2 \pi f}{F_s}}\right|^2\notag\\
	%&=\text{log} \sigma^2- \sum_{i=1}^{L} \text{log} \left| 1-A_i e^{\nicefrac{j 2 \pi (f_i-f)}{F_s}}\right|^2\notag\\
	&=\text{log} \sigma^2- \sum_{i=1}^{L} \text{log} \left|1+A_i^2-2A_i\text{cos}(\nicefrac{2 \pi (f_i-f)}{F_s})\right|
\end{align}
Therefore, the spectral density has a peak at $f_i$ Hz with power,
\begin{align}
	\text{log} P(f)|_{f=f_{i}} &=\text{log} \sigma^2-  2\text{log} \left|1-A_i\right|- \sum_{j=1,j\neq i}^{L} \text{log} \left|1+A_j^2-2A_j\text{cos}(\nicefrac{4 \pi (f_{j}-f_i)}{F_s})\right|
\end{align}
where the contribution of $p_i$ to the log power spectrum at $f_i$ Hz  is $-  2\text{log} \left|1-A_i\right|$. In linear scale, the power spectrum at $f_i$ Hz is proportional to $\nicefrac{1}{(1-A_i)^2}$. 

In summary, for a $L$ order LPC model, there are at most $\lceil\nicefrac{L}{2}\rceil$ dominant frequency peaks determined by the angles of the poles. On the other hand, power spectrum can be obtained directly from LPC coefficients $(a_1,a_2,\dots,a_L)$,
\begin{equation}
	\text{log} P(f)=\text{log} \sigma^2- 2\text{log} \left|1+\sum_{k=1}^{L}a_{k} e^{-\nicefrac{j 2 \pi f k}{F_s}} \right|
\end{equation}
 More detailed discussion can be found in \cite{takalo2005tutorial}.

 \subsection{Frequency-warped Linear Predictive Coding}\label{apdx_2}
 Frequency warping refers to a process that transforms a linear and uniformly spaced frequency scale, typically measured in Hertz (Hz), into a non-uniformly spaced frequency scale. This transformation is commonly applied to signal models and spectral representations in various fields such as signal processing, audio engineering, and telecommunications. 
 
 Frequency-warped linear predictive coding is a variation of LPC that modifies the spectral representation within the LPC framework by replacing the standard uniform frequency scale (unit delays $z^{-1}$) by first-order all-pass filters\cite{harma2000frequency,harma2000evaluation,roth2003frequency}. This enables it to become more sensitive to either high or low frequency components. In particular, for frequency-warped LPC the predictor transfer function in (\ref{eq:Hz}) becomes,
 \begin{equation}
 	H(z)=\frac{1}{1+\sum_{k=1}^{L}a_{k} D(z)^k}
 	\label{eq:Hz_warped}
 \end{equation}
 where,
 \begin{equation}
 	D(z)=\frac{z^{-1}-\lambda}{1-\lambda z^{-1}}.
 	\label{eq:D_warped}
 \end{equation}
 Here, $D(z)$ is a first order all-pass filter with warping coefficient $\lambda \in [-1,1]$.  Note that traditional LPC is a special case of frequency-warped LPC with $\lambda=0$. Frequency-warped LPC estimates non-uniform resolution spectral powers. The mapping from the natural frequency domain to the warped frequency domain can be obtained by the phase function of $D(z)$ which is given by\cite{harma2000frequency},
 \begin{equation}
 	f'=\frac{F_s}{2\pi} \mbox{tan}^{-1}\frac{(1-\lambda^2)\mbox{sin}(\nicefrac{2 \pi f}{F_s})}{(1+\lambda^2)\mbox{cos}(\nicefrac{2 \pi f}{F_s})-2\lambda}.
 	\label{eq:F_warped}
 \end{equation}
 In the z domain, this can be perceived as a bilinear transformation given by\cite{harma2000frequency},
 \begin{equation}
 	z^{-1}\rightarrow {z'}^{-1}=D(z)=\frac{z^{-1}-\lambda}{1-\lambda z^{-1}}.
 	\label{eq:Z_warped}
 \end{equation}
 and,
 \begin{equation}
 	{z'}^{-1}\rightarrow z^{-1}=\frac{{z'}^{-1}+\lambda}{1+\lambda {z'}^{-1}}.
 	\label{eq:Z_warped_inv}
 \end{equation}

 Thus, for positive values of $\lambda$, the resolution at low frequencies is increased and negative values of $\lambda$ yield a higher resolution at high frequencies.
 
 \subsection{Cepstrum coefficients}\label{apdx_3}
 For a stochastic model such as LPC, the cepstrum of the output process $(c_0,c_1,\dots,c_n,\dots)$ can be expressed in terms of the LPC coefficients $a_i$ or poles $p_k$ of the transfer function $H(z)$ \cite{boets2005clustering,kalpakis2001distance}. In particular, given the LPC coefficients $a=(a_1,a_2,\dots,a_{L})$, the cepstrum coefficients $c_n$ for the model can be calculated by,
 \begin{equation}
 	c_n = 
 	\begin{cases}
 		\text{log}\sigma^2 & \text{if } n =0 \\
 		-a_1 & \text{if } n =1 \\
 		-a_n-\sum_{m=1}^{n-1} (1-\frac{m}{n}) a_m c_{n-m}& \text{if } 1 < n\leq L \\
 		-\sum_{m=1}^{L} (1-\frac{m}{n}) a_m c_{n-m}& \text{if } L<n \\
 	\end{cases}
 \end{equation}
 
 Alternatively, given the poles $p_k$ of the model, $c_n$ can be calculated by,
 \begin{equation}
 	c_n = 
 	\begin{cases}
 		\text{log}\sigma^2 & \text{if } n =0 \\
 		\frac{1}{n}\sum_{m=1}^{L} p_m^{n}& \text{if } n> 0 \\
 	\end{cases}
 \end{equation}
 Conversely, we can obtain LPC coefficients $a_i$ from the cepstrum coefficients by,
 \begin{equation}
 	a_i=\begin{cases}
 		-c_1 &\text{if } i=1\\
 		-c_i -\sum_{m=1}^{i-1} \big(1-\frac{m}{i}\big) a_m c_{i-m} &\text{if } 1<i\leq L\\
 	\end{cases}
 \end{equation}
 The euclidean distance metric between two LPC models $a$ and $a'$ can be defined by\cite{lauwers2017time},
 \begin{equation}
 	d_{\text{CEP}\infty}(a,a'):=\sqrt{ (c_0-c'_0)^2+\sum_{i=1}^{\infty} i (c_i-c'_i)^2}
 \end{equation}
 However, this creates a latent space with non-finite dimensions. Therefore, we utilize first $M$ coefficients,
 \begin{equation}
 	d_{\text{CEP}M}(a,a'):=\sqrt{(c_0-c'_0)^2+\sum_{i=1}^{M} i (c_i-c'_i)^2}
 \end{equation}
 It can be readily seen that $d_{\text{cep}M}(.)$ provides a lower bound for $d_{\text{cep}\infty}(.)$. Finally, we create our latent space such that $d_{\text{cep}M}(.)$ naturally holds,
 \begin{equation}
 	F_2(a):=(c_0,c_1,\sqrt{2}c_2,\dots,\sqrt{M}c_M)
 \end{equation}
 
 Note that another possible way to generate a latent space could be to use the poles $p_k$ of the transfer function $H(z)$ in (\ref{e3}). However, the distance between two LPC models in this space is not Euclidean. Specifically, the squared distance between two LPC models in this space with poles $p$ and $p'$ is can be defined by \cite{martin2000metric},
 \begin{equation}
 	d^2_{POLE}(a,a'):=\text{ln} \frac{\prod_{i} \prod_{j} (1- p_i {p'_j}^{*}) (1- p_j {p'_i}^{*})}{\prod_{i=1}^{L} \prod_{j} (1- p_i {p_j}^{*}) \prod_{i} \prod_{j} (1- p'_i {p'_j}^{*})} 
 \end{equation} 
 where $i,j\in\{ 1,2,\dots,L\}$. This makes it harder to use traditional clustering tools such as k-means in this latent space. Instead, one can consider k-medoids with the above distance metric. However, such algorithms have higher complexity which is not suitable for large-scale datasets. Most importantly, an equivalence between $d_{CEP\infty}(.)$ and $d_{POLE}(.)$ can be shown for $c_n; i>0$ \cite{martin2000metric}.
 %\newpage
 
 \subsection{BERT architecture and parameters}\label{apdx_4}

In our work, we utilized BERT model tailored for masked language modeling (MLM) tasks during self-supervised learning. The BERT model's configuration includes a hidden size of 256, with 6 transformer layers, each containing a single attention head. We employed a relative key position embedding strategy, which allows the model to better capture the relative positioning of tokens within a sequence. Additionally, we enabled the output of hidden states, which were utilized for the downstream task. Model details are given in Table \ref{tab:my-table}.
\newpage
  \begin{table}[ht]
 	\centering
 	\begin{tabular}{|l|l|l|}
 		\hline
 		\textbf{Layer} & \textbf{Output Shape} & \textbf{Param \#} \\ \hline
 		BertModel: 1-1 & [1, 64, 256] & -- \\ \hline
 		\quad BertEmbeddings: 2-1 & [1, 64, 256] & 32,768 \\ \cline{2-3}
 		\quad \quad Embedding: 3-1 & [1, 64, 256] & 3,328 \\
 		\quad \quad Embedding: 3-2 & [1, 64, 256] & 512 \\
 		\quad \quad LayerNorm: 3-3 & [1, 64, 256] & 512 \\
 		\quad \quad Dropout: 3-4 & [1, 64, 256] & -- \\ \hline
 		\quad BertEncoder: 2-2 & [1, 64, 256] & -- \\
 		\quad \quad ModuleList: 3-5 & -- & 11,237,376 \\ \hline
 		BertOnlyMLMHead: 1-2 & [1, 64, 69] & -- \\ \hline
 		\quad BertLMPredictionHead: 2-3 & [1, 64, 69] & -- \\ \cline{2-3}
 		\quad \quad BertPredictionHeadTransform: 3-6 & [1, 64, 256] & 66,304 \\
 		\quad \quad Linear: 3-7 & [1, 128, 69] & 17,733 \\ \hline
 	\end{tabular}%
 	\caption{Detailed architecture and parameters for BERT model utilized in this study.}
 	\label{tab:my-table}
 \end{table}
 
\end{document}